\documentclass[review]{elsarticle}

\journal{Pattern Recognition}

\usepackage{lineno,hyperref}%

\usepackage{pstricks}
\usepackage{pst-eps}
\usepackage{amsthm,amsmath,mathrsfs,amssymb}
\usepackage{epstopdf}
\usepackage{bbm}
\usepackage{color}
\usepackage{comment}
\usepackage{multirow}
\usepackage{bm}
\usepackage{float}
\usepackage[tight]{subfigure}
\usepackage{graphicx,color,url}
\usepackage{subfigure}

\usepackage{keyval}
\usepackage{algorithm}
\usepackage{algorithmic}
\usepackage{tabularx,booktabs}
\usepackage{extarrows}
\usepackage{threeparttable}
\usepackage{longtable}
\usepackage{caption}
\usepackage{float}

\newsavebox{\tempbox}

\newtheorem{theorem}{Theorem}

\usepackage{tikz}
\usepackage{etoolbox}
\usepackage{enumitem}
\newcommand*{\circled}[1]{\lower.7ex\hbox{\tikz\draw (0pt, 0pt)%
		circle (.5em) node {\makebox[1em][c]{\small #1}};}}
\robustify{\circled}

\begin{document}

\begin{frontmatter}

\title{Multi-Instance Learning by Utilizing Structural Relationship among Instances}

\author{Yangling Ma}
\author{Zhouwang Yang\corref{mycorrespondingauthor}}
\cortext[mycorrespondingauthor]{Corresponding author}
\ead{yangzw@ustc.edu.cn}
\address{University of Science and Technology of China, Hefei 230026, P. R. China}




\begin{abstract}
	Multi-Instance Learning (MIL) aims to learn the mapping between a bag of instances and the bag-level label. Therefore, the relationships among instances are very important for learning the mapping. In this paper, we propose an MIL algorithm based on a graph built by structural relationship among instances within a bag. Then, Graph Convolutional Network (GCN) and the graph-attention mechanism are used to learn bag-embedding. In the task of medical image classification, our GCN-based MIL algorithm makes full use of the structural relationships among patches (instances) in an original image space domain, and experimental results verify that our method is more suitable for handling medical high-resolution images. We also verify experimentally that the proposed method achieves better results than previous methods on ﬁve benchmark MIL datasets and four medical image datasets.

\end{abstract}


\end{frontmatter}


\section{Introduction}\label{sec:intro}
In many computer vision problems, such as image classification, an image clearly represents a category. However, in many real-world applications, multiple instances are observed and only one label is given to the whole of these instances. This scenario is commonly named Multi-Instance Learning (MIL)~\cite{zhou2009multi}. The problem is particularly evident in high-resolution medical image analyses~\cite{combalia2018monte}, e.g., mammography and whole-slide tissue images, where a single label (benign/malignant) or region of interest (ROI) is often used to roughly describe image categories. A common solution is to divide the high-resolution medical image into patches and then use patch-level annotations to train a supervised classifier~\cite{wang2018deep}. However, for high-resolution medical images, obtaining patch-level labels is labor-intensive and time-consuming~\cite{xu2019camel}. Therefore, applying weakly-supervised MIL, which only requires labels from whole images rather than patch-level labels, on high-resolution medical image classification problems is a better solution~\cite{hou2016patch,combalia2018monte}.  

Multi-Instance Learning (MIL) treats a bag that is composed of several instances as the entirety, where only the bag is labeled and the instances are unlabeled~\cite{zhou2009multi}. Therefore, for multi-instance learning methods, an important challenge is to study relationships among instances in an entire bag, such as similarity relationships among instances~\cite{zhou2009multi}. For processing the relationships among instances, one type of method based on the independent distribution among instances establishes a many-to-one mapping function, and the other type of method based on the non-independent distribution among instances also establishes a many-to-one mapping function.

Different methods are proposed based on the independent distribution among instances, such as using the similarity between bags~\cite{cheplygina2015label}, embedding an instance into a compact low-dimensional representation and then passing from the low-dimensional representation to the bag-level classifier~\cite{andrews2003support,chen2006miles}, as well as combining with the response of the instance-level classifier~\cite{raykar2008bayesian,zhang2006adapting}. Only the last method can provide interpretable results, but the instance-level accuracy of this method is still low~\cite{kandemir2015computer}. Recently, an Deep Neural Network (DNN) has been applied to MIL~\cite{wang2018revisiting,ilse2018attention}. MIL algorithms based on DNNs outperform existing shallow algorithms~\cite{wang2018revisiting}, where the basic idea is to add a pooling operation in the process of learning bag-embedding in DNN. In~\cite{ilse2018attention}, an attention mechanism for instance concentration is introduced instead of an untrainable pooling operation~\cite{wang2018revisiting}. Trainable attention weights on instances provide additional information about the contribution of each instance to the final decision~\cite{ilse2018attention}. This approach~\cite{ilse2018attention} provides a reasonable explanation for final bag classification but still treats instances as independent. 

For some special MIL problems, such as high-resolution medical image classification, the patches in the original image space domain are non-independent. Xu et al.~\cite{xu2014weakly} combine the similarity relationships among patches with the MIL method for medical image segmentation and classification. Jia et al.~\cite{jia2017constrained} use the context relationships between patches as constraints for image segmentation. With the development of graph networks in recent years, in~\cite{tu2019multiple}, graphs are established based on similarity among instances in the feature value domain, and they learn bag-embedding by using a Graph Neural Network (GNN). Although this method improves the accuracy and considers similarity among instances, it dose not directly use the structural relationships among patches in the original image space domain for medical image classification~\cite{tu2019multiple}. 

In this paper, we propose a new approach to incorporate structural relationships among instances into the MIL approach. We build a graph by using the structural relationships among instances and learn bag-embedding based on graph convolutional networks (GCNs) and the graph-attention mechanism. Starting from the basic theorem of continuous function on invariance of variable arrangement, we provide a verification theorem for a bag-scoring function based on GCN, which includes three steps: instance conversion to a low-dimensional embedding, an aggregation function based on graph convolution and a graph-attention mechanism, and final transformation into bag probability. Finally, we propose an instance-embedding formula based on a graph-attention mechanism to replace the original method~\cite{ilse2018attention,tu2019multiple}. Notably, attention weights allow us to find key instances that can be further used to obtain ROI. As shown in experimental results, our algorithm outperforms other methods on five MIL benchmark datasets and four medical image datasets. In addition, our method is more suitable for handling high-resolution medical image classification. Overall, this paper makes three main contributions:

\begin{itemize}
	\item First, we build a graph utilizing the structural relationships among instances in the image space domain,  and use a GCN based on this graph to extract better feature representations of instances.
	
	\item Second, an end-to-end MIL algorithm with a new graph-attention mechanism is proposed, which significantly improves the accuracy of the MIL algorithm.
	\item Third, for high-resolution medical image classification tasks, we transform them into multi-instance learning problems, and achieve the best available results by using our proposed method.
\end{itemize}

The rest of this article is organized as follows. Section~\ref{sec:relate} briefly reviews previous studies on MIL. In Section~\ref{sec:intro}, we propose end-to-end MIL networks. Our experimental results and some discussions of experimental setups are presented in Section~\ref{sec:expe}. Finally, in Section~\ref{sec:con}, we conclude the article with some future studies.

\section{Related Work}\label{sec:relate}
 Different methods have been proposed based on the independent distribution among instances, such as using the similarity between bags~\cite{cheplygina2015label}. In traditional MIL methods~\cite{cheplygina2015label,chen2006miles,raykar2008bayesian}, instances are represented by precomputed features, and additional feature extraction is rarely required. However, these traditional methods obtain feature representations of instances through neural networks, then learn bag-embedding through pooling operations, and finally obtain bag scores through a full-connected layer~\cite{wang2018revisiting}. The MIL seems well suited for medical image analysis because deep networks are computationally unfeasible for processing whole-slide images consisting of billions of pixels~\cite{combalia2018monte}. In addition, pixel-level annotations are difficult to obtain in the medical field. Therefore, it is natural to segment medical image into smaller patches, and then regard these patches as instances~\cite{hou2016patch}. Common methods of patch segmentation are usually the grid-sampling strategy and uniform-sampling strategy~\cite{hou2016patch}. However, these patches are not representative. Aiming at the problem of high-resolution image classification, Combalia $\&$ Vilaplana~\cite{combalia2018monte} propose a patch-sampling strategy based on a sequential Monte-Carlo method, and iterate to get representative patches. In our method, we use the structural relationships among patches in the original image space domain to establish a graph.

 Different methods have been proposed to incorporate relationships among instances in MIL, such as using graphs~\cite{zhou2009multi} and attention mechanisms~\cite{ilse2018attention}. Previous MIL works based on graphs are mainly proposed by Zhou et al., Zhang et al. and Tu et al. in~\cite{zhou2009multi,zhang2011multiple,tu2019multiple}. Most MIL algorithms treat instances in each bag as independent and identically distributed samples~\cite{zhou2007relation}. This strategy ignores the structural relationships among instances in each bag. This assumption is invalid under actual conditions, such as patches in high-resolution image classification~\cite{zhou2009multi}. Experimental results in~\cite{zhou2009multi} show that the method of constructing a graph for each bag and kernel learning of the graph outperforms previous methods. Zhang et al.~\cite{zhang2011multiple} use the relationships among bags or instances to build a graph. Next, they add a graph regularization term based on the given graph of ordinary MIL loss that enforces smoothness on the soft labels of nodes in the given graph. A graph is built through the similarities among instances in the feature value domain, and then they learn the bag-embedding by graph neural networks (GNNs)~\cite{tu2019multiple}. 
 
 In addition, for tasks with structured relational data, such as high-resolution image classification, where each image is a bag and patches are instances, it is unnatural to treat model input as irrelevant instances. In our method, a graph of a bag is established by using the structural relationships among patches in the image space domain. In addition, in order to get a better bag-embedding, a attention mechanism, which is based on independent instances, acts as a kind of permutation invariant aggregation operator based on neural networks~\cite{tu2019multiple,ilse2018attention}. In our method, a new graph-attention mechanism is proposed by introducing the structural relationships among instances.

\section{Methodology}\label{sec:method}

In this section, we detail the mathematical basis of GCN-based MIL and our algorithm.

\subsection{Mathematical description of GCN-based MIL}\label{sec:method:mil}

Multi-instance learning (MIL), as a weakly supervised learning algorithm, processes weakly annotated data, where in each data sample (often called a bag) has multiple instances but merely one label~\cite{zhou2009multi}. To some extent, the MIL can be expressed by a formula as a supervised learning task, a bag as input, and a bag-level label as target. Suppose a set of bags is defined as $[X^1,X^2,\dots,X^N]$, and each bag $X^i$ contains $K$ instances $[x^i_1,x^i_2,\dots,x^i_K]$. The purpose of the MIL is to learn a mapping function from $N$ bags to corresponding labels $[Y^1,Y^2,\dots,Y^N]$. Assuming that the mapping function is $S(X)$, for a typical two-category MIL problem, if the bag contains an instance of a positive class, this bag is a positive sample; otherwise, this bag is a negative sample (as shown in Equation~\eqref{eqn:1}). These assumptions indicate that the mapping function $S(X)$ must be permutation invariant~\cite{ilse2018attention}.
\begin{equation}
\begin{split}
Y^i  &= \begin{cases}            0, &\sum_k y^i_k = 0, \\            1, &otherwise,         \end{cases}
\end{split}
\label{eqn:1}
\end{equation}
where $y^i_k$ represents the label corresponding to each instance $x^i_k$.


For MIL benchmark problems, they have extracted initial feature representations of instances by using other strategies, so the input bag $X^i$ is converted into a graph $(A^i, V^i)$ through the similarities among instances in the eigenvalue domain, wherein the adjacent matrix $A^i$ is established in a manner similar to the heuristic strategy used in~\cite{zhou2009multi} (Equation~\eqref{eqn:2}). For image classification tasks, initial feature representations of instances $x_k^i$ are extracted by using CNN since CNN is one of the best feature extractors~\cite{ma2020pcfnet:}. As shown in Equation~\eqref{eqn:3}, the adjacent matrix $A^i$ is established using structural relationships (1-neighborhood) among patches in the original image space domain. In addition, the dimension of this initial feature vector is $F_1$, and the transformation function of a trainable CNN or other untrainable strategies is $f$. The specific CNN structures are given in experiments. 

\begin{equation}
	\begin{split}
		A^i_{mn}  &= \begin{cases}            1, &dist(x^i_m,x^i_n) > 0.5, \\            0, &otherwise,         \end{cases}
	\end{split}
	\label{eqn:2}
\end{equation}
where $x_m^ i$ and $x_n^ i$ represent the m and n instances in bag $X^i$, and $dist(x_m^i,x_n^i)$ represent the similarity between $x_m^ i$ and $x_n^ i$. The measuring function used in this paper is the cosine distance.
\begin{equation}
	\begin{split}
		A^i_{mn}  &= \begin{cases}            1, &x^i_m,x^i_n\ is\ adjacent\ to\ an\ edge\ or\ vertex, \\            0, &otherwise,         \end{cases}
	\end{split}
	\label{eqn:3}
\end{equation}
where $x_m^ i$ and $x_n^ i$ represent the m and n instances (patches) in a bag (image) $X^i$.

It should be noted that the transformation function $M(X)$ of the GCN-based MIL can be expressed as the score function of bags. In order to facilitate theoretical analysis, only the single-layer graph convolution is considered. Since the score function $S(X)$ of bags features permutation invariance, we only need to prove that M(X) also demonstrates permutation invariance. Referring to~\cite{zaheer2017deep}, the following theorem is obtained from the basic theorem of continuous function for the invariance of variable arrangement. 

\begin{theorem}\label{sec:method:milgcn:thm1}
	If $M(X)$ is the transformation function of the GCN-based MIL, then $M(X)$ remains unchanged for any permutation transformation $\pi$ acting on $X$. This is true where $f$ and $\sigma$ are both reversible continuous functions; $W\in R^{F_1 \times F_2}$ represents the parameter variables of GCN to learn; and $N(k)$, $\alpha_k$, and $d_k$ represent the 1-neighborhood of the instance, the attention weight of the instance, and the degree of the instance. $M(X)$ is expressed as follows:
	\begin{equation}
	\begin{split}
	M(X) &= M(x_1,x_2,\dots,x_K) \\
	&= \sigma\left(\sum_{k=1}^K{\alpha_k\frac{\sum_{j\in{N(k)}}{f(x_j)^TW}}{d_k}}\right)
	\end{split}
	\label{eqn:thm1}
	\end{equation}
\end{theorem}
\begin{proof}
	
	Since any permutation transformation $\pi$ acting on $X$ does not change set $N(k)$ for the neighborhood nodes of the node $x_k$, the derivation process is as follows:
	\begin{equation}
	\begin{split}
	M(\pi(X)) &= M(x_{\pi(1)},x_{\pi(2)},\dots,x_{\pi(K)}) \\
	&= \sigma\left(\sum_{k=1}^K{\alpha_{\pi(k)}z_{\pi(k)}}\right) \\
	&=  \sigma\left(\sum_{k=1}^K{\alpha_{\pi(k)}\frac{\sum_{j\in{N(\pi(k))}}{f(x_j)^TW}}{d_{\pi(k)}}}\right) \\
	&=  \sigma\left(\sum_{k=1}^K{\alpha_k\frac{\sum_{j\in{N(k)}}{f(x_j)^TW}}{d_k}}\right) \\
	&= M(x_1,x_2,\dots,x_K) \\
	&= M(X)
	\end{split}
	\label{eqn:thm2}
	\end{equation}
\end{proof}

It is known from Theorem~\ref{sec:method:milgcn:thm1} that $M(X)$ exhibits permutation invariance, so $M(X)$ can be expressed as the score function $S(X)$ of a bag.

\subsection{Proposed algorithm}\label{sec:method:milgcn}
In this section, we detail the new MIL algorithm in main parts. First, the whole framework of the algorithm is introduced, and then the two most important parts of the algorithm, the graph convolution network and the graph-attention mechanism, are introduced respectively.
\subsubsection{The framework of the algorithm}
\begin{figure}[H]
	\centering
	\includegraphics[width=0.9\textwidth]{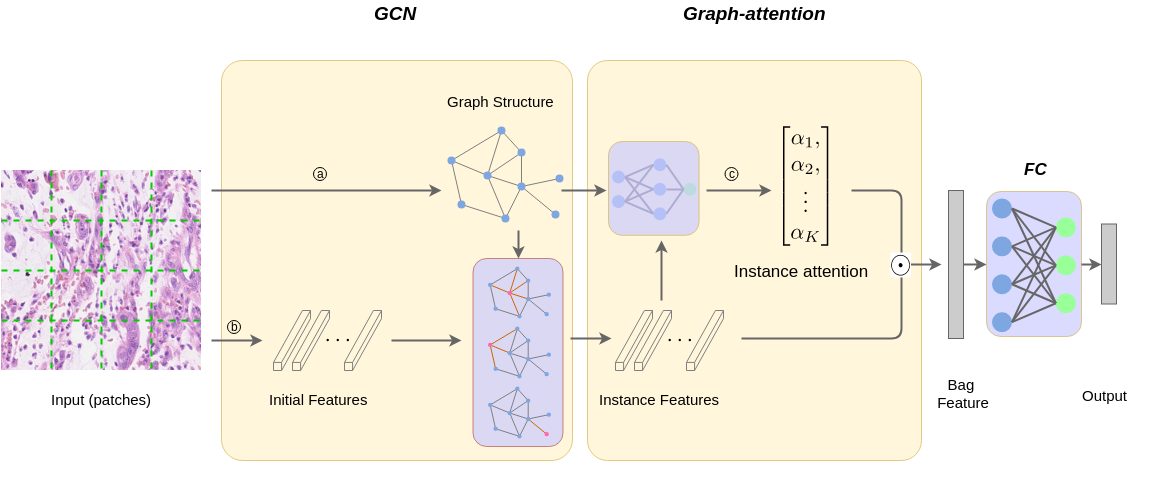}
	\caption{GCN-based MIL framework overview. \circled{a}: This process uses Equation~\eqref{eqn:2} or~\eqref{eqn:3} to establish the adjacent matrix in the graph. \circled{b}: This process uses a convolutional neural network to extract initial feature vectors from the instance. \circled{c}: This process incorporates a graph structure into an attention mechanism to calculate the weighting factor $\alpha_k$ for each instance (patch) $x_k$.}
	\label{fig:1}
\end{figure}

As shown in Figure~\ref{fig:1}, the framework of this new MIL algorithm mainly consists of three primary steps. Firstly, we use a GCN based on graph and initial feature representations of instances to obtain mingled representations $z_k^i$ of instances. Then, we incorporate the graph structure into the attention mechanism to create the graph-attention mechanism, which calculates the weighting factor $\alpha_k^i$ for each instance $x^i_k$. We combine the instances with attention weights to obtain the feature representation $Z^i$ of the bag $X^i$. Finally, we convert the feature representation of the bag into a class score for the bag through a fully-connected layer, and the transformation function is $\sigma$.

The above process introduces the forward propagation process of the method. For the multi-instance learning method based on graph convolutional networks, we use the back propagation algorithm and the gradient descent method with learning rate for training to obtain all parameter values in the framework. 

\subsubsection{The graph convolution network}

Some studies have shown that better performance is achieved in classification and regression tasks by considering the relationships among instances in bags~\cite{tu2019multiple,zhou2009multi}. As shown in~\cite{tu2019multiple}, if a bag is regarded as a graph, the fraction learning of a bag is consistent with graph classification, so they establish the adjacent matrix $A$ in the graph by using the similarity among instances in the feature value domain. They regard the set of instances $[x^i_1,x^i_2,\dots,x^i_K]$ as the vertex set $V^i$ in the graph, and finally get the category score of a bag by combining the GNN. In recent years, with the in-depth study of graph-structure data learning tasks, it was found in~\cite{kipf2016semi} that GCN simultaneously carries out end-to-end learning of node feature information and structural information, and GCN is suitable for nodes and graphs of arbitrary topology. In this paper, we establish the adjacent matrix $A$ in the graph by using structural relationships among instances in the original image space domain, and we use GCN to obtain feature representations of instances.

The input of the graph convolution network in our method is made up of initial feature representations of instances and a graph. In addition, for image classification tasks, we use Equation~\eqref{eqn:3} to establish adjacent matrix $A$ in a graph and use a CNN (like ResNet) to extract initial feature representations of instances. For MIL benchmark problems, we use 
Equation~\eqref{eqn:2} to establish adjacent matrix $A$ in a graph and initial feature representations of instances have been provided. Generally speaking, we use the GCN to obtain mingled representations $z_k^i$ of instances, the dimension of $z_k^i$  is $F_2$. The graph convolution formula used in GCN is as shown in Equation~\eqref{eqn:4}. 
\begin{equation}
\begin{split}
z^i_k &= \frac{\sum_{j \in{N^i(k)}}{{f(x^i_j)}^TW}}{d^i_k}
\end{split}
\label{eqn:4}
\end{equation}
where $N^i(k)$ represents the set of neighborhood nodes for the $k$ instance in bag $X^i$, $d^i_k=\sum_{j=1}^{K}A^i_{kj}$ represents the degree of the $k$ instance in bag $X^i$, and $W \in R^{F_1 \times F_2}$ represents the parameter variables to learn.



\subsubsection{The graph-attention mechanism}\label{sec:method:gam}
Attention mechanism operators mentioned in previous works~\cite{tu2019multiple,ilse2018attention} have a clear disadvantage, namely, they are employed on feature representation of single instance. They use a weighted average of instances (low-dimensional embeddings), where weights are determined by a neural network. In order to use the structural relationships among instances, we propose a new graph-attention mechanism based on the given graph and feature representations of instances to obtain weights of instances. Let $[z^i_1,\dots, z^i_K]$,  $[d^i_1,\dots, d^i_K]$, and $[N^i(1), \dots, N^i(K)]$ be feature representations of K instances, degrees of K instances, and 1-neighborhoods of K instances, respectively. Then, we propose the following graph-attention mechanism:
\begin{equation}
	\begin{split}
		\alpha_k^i &= softmax\left(u^Ttanh\left(\frac{\sum_{j\in{N^i(k)}}{Vz^i_j}}{d^i_k}\right)\right)
	\end{split}
	\label{eqn:5}
\end{equation}

\begin{equation}
	\begin{split}
		Z^i &= \sum_{k=1}^K{\alpha_k^iz_k^i}
	\end{split}
	\label{eqn:51}
\end{equation}
where $u \in R^{L\times 1},V\in R^{L\times F_2}$ represents the parameter variables to learn. The $softmax$ function is really just a normalization function that normalizes a vector to 1 in a special way. $Z^i$ is the feature representation of bag, and the $tanh$ function is an activation function.

Interestingly, the proposed graph-attention mechanism corresponds to a version of the attention mechanism~\cite{ilse2018attention}. All instances are sequentially dependent or independent in the typical attention mechanism, while here we assume that the relationships among all instances are represented by a graph.  Therefore, a naturally arising question is whether the attention mechanism could work based on a graph of instances. We will address this issue in the experiments.

\section{Experiments}\label{sec:expe}

In this section, we introduce some datasets and experimentally verify the performance of the GCN-based MIL algorithm proposed in this paper. We want to validate three research questions in the experiment: (i) our method achieves the best performance compared to the best MIL method, and (ii) we explain the effectiveness of the graph attention mechanism and the graph convolutional network for multi-instance learning methods through experiments, (iii) our method is more suitable for high-resolution medical image classification tasks. 

\subsection{Datasets}\label{sec:expe:dataset}

\textbf{MIL benchmark datasets~\cite{ilse2018attention}} Five datasets contain pre-computed features and only a small number of instances and bags. As shown in Table~\ref{fig:2}, MUSK1 and MUSK2 are datasets for drug activity prediction, and FOX, TIGER, and ELEPHANT are image datasets. The total number of bags, the total number of instances, the number of positive and negative samples, and the feature vector dimensions they contain are shown in Table~\ref{fig:2}, respectively.  To obtain a fair comparison, we use a common evaluation methods, i.e., 10-fold cross validation, and each experiment is repeated five times~\cite{ilse2018attention}.

\begin{table}[t]
	\scriptsize
	\centering
	\caption{Five MIL benchmark datasets.}
	\begin{tabular}{|l|r|r|r|r|r|r|}
		\hline
		\multirow{2}{*}{Dataset} & \multirow{2}{*}{Attributes} & \multicolumn{3}{c|}{Bags} & \multirow{2}{*}{Instances} & \multirow{2}{*}{Average Bag Size} \\ \cline{3-5}
		& & Positive & Negative & Total & & \\ \hline
		\multicolumn{7}{|c|}{Drug Activity Prediction} \\ \hline
		Musk1 & 166 & 47 & 45 & 92 & 476 & 5.17 \\ \hline
		Musk2 & 166 & 39 & 63 & 102 & 6589 & 64.69 \\ \hline
		\multicolumn{7}{|c|}{Content-based image retrieval and classification} \\ \hline
		Tiger & 230 & 100 & 100 & 200 & 1391 & 6.96 \\ \hline
		Elephant & 230 & 100 & 100 & 200 & 1220 & 6.10 \\ \hline
		Fox & 320 & 100 & 100 & 200 & 1320 & 6.60 \\ \hline
	\end{tabular}
	\label{fig:2}
\end{table}

\textbf{Breast cancer dataset~\cite{ilse2018attention}.} This dataset consists of 58 weakly labeled 896$\times$768 Hematoxylin and Eosin ($H\&E$) images. If an image contains breast cancer cells, it is marked as malignant; otherwise, it is benign. We divide each image into $32\times32$ patches and move each patch in steps of 32.

\textbf{Colon cancer dataset~\cite{ilse2018attention}.} This dataset includes a total of 100 $H\&E$ images; the resolution of each image is $500\times500$. The images are derived from various tissues from normal and malignant areas. For each image, most of the nucleus of each cell is labeled. We divide each image into $27\times27$ patches and move each patch in steps of 27. In addition, once one instance (patch) contains one or more epithelial nuclei, a positive bag marker is given; otherwise, a negative bag maker is given. Labeling epithelial cells is highly correlated from a clinical perspective because colon cancer originates from epithelial cells.

\textbf{ICIAR dataset~\cite{wang2018classification}.} The ICIAR dataset consists of 400 mammographic images, and there are four categories: normal, benign, invasive, and carcinoma in situ. For each class, there are 100 different Hematoxylin and Eosin ($H\&E$) stained images with a resolution of $2048\times1536$. For each experiment, 320 images are used for training, and 80 images are used for testing. These patches have been previously cropped from the whole-slide tissue image that is labeled, so from a MIL perspective, the labels patches (instances) we extracted from one image (one bag) are consistent with the image label. We divide each image into $224\times224$ patches and move each patch in steps of 200.

\textbf{IDRiD dataset~\cite{idrid}.} The IDRiD dataset was published as a challenge dataset in ISBI 2018. One purpose of this challenge was to evaluate the automatic disease grading algorithm for diabetic retinopathy (DR) and diabetic macular edema (DME) by using fundus images. The training data contains 413 images, and the test data contains 103 images. The resolution of each image is $4288\times2848$.  We divide each image into $224\times224$ patches and move each patch in steps of 200.

\subsection{Experimental settings}\label{sec:expe:model}
For experiments on five MIL benchmark datasets, the basic architecture of the deep network we used is the same as~\cite{wang2018revisiting}, but the fully-connected layer, except the last layer, is replaced by the graph convolutional layer. In addition, the super parameter $L$ in the graph-attention mechanism is 64. For experiments on the breast cancer dataset and colon cancer dataset, the basic architecture of the deep network we used is the same as~\cite{ilse2018attention}, but the fully-connected layer except the last layer is replaced by the graph convolutional layer. In addition, the super parameter $L$ in the graph-attention mechanism is 64.  For experiments on the ICIAR dataset and IDRiD dataset, we first extract the feature vector of the patch by using ResNet18 without the fully-connected layer ($f$) and then connect a graph convolution layer and a graph-attention mechanism ($L=128$) to obtain bag-embedding. Finally, we connect a fully-connected layer to obtain the category score of a bag.

For each layer of networks used in this paper, we use the Xavier initialization strategy to initialize parameters. For parameters in ResNet18, we use a pretrained model on ImageNet to initialize. For each experiment we set the batch size to one and the largest epoch to 200. Finally, we take the model with the smallest loss function on the validation set to predict unknown bags. All models are trained using the Adam optimization algorithm. All experiments are performed on a single machine with GPU TITAN X (Pascal) and RAM 32G.

\subsection{Experimental results and analyses}\label{sec:expe:bench}
In the first experiment, our purpose is to verify whether our method surpasses other MIL methods on five MIL benchmark datasets. Each dataset is divided into ten folds, and we use nine folds for training and one fold for testing. For the five MIL benchmark datasets, the instances in the bag have similarities in the feature value domain. Therefore, in these experiments, we used Equation~\eqref{eqn:2} to create the adjacent matrix $A$ in the graph.

\begin{table}
	\footnotesize
	\centering
	\caption{Results on classic MIL datasets. Experiments were run five times, and the average of the classification accuracy ($\pm$ the standard error of the mean) is reported.}
	\resizebox{\textwidth}{!}{
		\begin{tabular}{llllll}
			\hline\noalign{\smallskip}
			Method & MUSK1 & MUSK2 & FOX & TIGER & ELEPHANT\\
			\noalign{\smallskip}\hline\noalign{\smallskip}
			mi-SVM~\cite{andrews2003support} & $0.874\pm N/A$ & $0.836\pm N/A$ & $0.582\pm N/A$ & $0.784\pm N/A$ & $0.822\pm N/A$\\
			MI-SVM~\cite{andrews2003support} & $0.779\pm N/A$ & $0.843\pm N/A$ & $0.578\pm N/A$ & $0.840\pm N/A$ & $0.843\pm N/A$  \\
			MI-Kernel~\cite{gartner2002multi} & $0.880\pm 0.031$ & $0.893\pm 0.015$ & $0.603\pm 0.028$ & $0.842\pm 0.01$ & $0.843\pm 0.016$ \\
			EM-DD~\cite{zhang2002dd} & $0.849\pm 0.044$ & $0.869\pm 0.048$ & $0.609\pm 0.045$ & $0.730\pm 0.043$ & $0.771\pm 0.043$ \\
			mi-Graph~\cite{zhou2009multi} & $0.889\pm 0.033$ & $0.903\pm 0.039$ & $0.620\pm 0.044$ & $0.860\pm 0.037$ & $0.869\pm 0.035$ \\
			miVLAD~\cite{wei2016scalable} & $0.871\pm 0.043$ & $0.872\pm 0.042$ & $0.620\pm 0.044$ & $0.811\pm 0.039$ & $0.850\pm 0.036$ \\
			miFV~\cite{wei2016scalable} & $0.909\pm 0.04$ & $0.884\pm 0.042$ & $0.621\pm 0.049$ & $0.813\pm 0.037$ & $0.852\pm 0.036$ \\
			\noalign{\smallskip}\hline
			mi-Net~\cite{wang2018revisiting} & $0.889\pm0.039$ & $0.858\pm 0.049$ & $0.613\pm 0.035$ & $0.824\pm 0.034$ & $0.858\pm 0.037$ \\
			MI-Net~\cite{wang2018revisiting} & $0.887\pm0.041$ & $0.859\pm0.046$ & $0.622\pm0.038$ & $0.830\pm0.032$ & $0.862\pm0.034$ \\
			MI-Net with DS~\cite{wang2018revisiting} & $0.894\pm0.042$ & $0.874\pm0.043$ & $0.630\pm0.037$ & $0.845\pm0.039$ & $0.872\pm0.032$ \\
			DNN+Attention~\cite{ilse2018attention} & $0.892\pm 0.04$ & $0.858\pm 0.048$ & $0.615\pm0.043$ & $0.839\pm0.022$ & $0.868\pm0.022$ \\
			DNN+Gate-Attention~\cite{ilse2018attention} & $0.900\pm0.05$ & $0.863\pm0.042$ & $0.603\pm0.029$ & $0.845\pm0.018$ & $0.857\pm0.027$ \\
			GNN+Attention~\cite{tu2019multiple} & $0.917\pm0.048$ & $0.892\pm0.011$ & $0.679\pm0.007$ & $0.876\pm0.015$ & $0.903\pm 0.010$ \\
			\noalign{\smallskip}\hline
			ours & $\bf{0.927\pm0.013}$ & $\bf{0.915\pm0.009}$ & $\bf{0.690\pm0.017}$ & $\bf{0.889\pm0.008}$ & $\bf{0.909\pm0.005}$ \\
			\noalign{\smallskip}\hline
	\end{tabular}}
	\label{tab:1}
\end{table}

As shown in Table~\ref{tab:1}, our method outperforms the other MIL algorithms, which include DNN-based MIL algorithms~\cite{wang2018revisiting,ilse2018attention}, traditional non-DNN MIL algorithms~\cite{andrews2003support,zhou2009multi,wei2016scalable,zhang2002dd,gartner2002multi}, and GNN-based MIL algorithms~\cite{tu2019multiple}, on five benchmark datasets.  In addition, the last line of Table~\ref{tab:1} verifies that the GCN-based MIL algorithm combined with the graph-attention mechanism is more suitable for MIL problems.

\begin{table}
	\footnotesize
	\centering
	\caption{Comparision with baselines. Experiments were run five times, and the average classification accuracy ($\pm$ the standard error of the mean) is reported.}
	\resizebox{\textwidth}{!}{
		\begin{tabular}{llllll}
			\hline\noalign{\smallskip}
			Method & MUSK1 & MUSK2 & FOX & TIGER & ELEPHANT\\
			\noalign{\smallskip}\hline\noalign{\smallskip}
			DNN+Attention(baseline) & $0.892\pm 0.04$ & $0.858\pm 0.048$ & $0.615\pm0.043$ & $0.839\pm0.022$ & $0.868\pm0.022$ \\
			GNN+Attention & $0.917\pm0.048$ & $0.892\pm0.011$ & $0.679\pm0.007$ & $0.876\pm0.015$ & $0.903\pm 0.010$ \\
			GCN+Attention & $0.923\pm0.071$ & $0.909\pm0.008$ & $0.683\pm0.023$ & $0.881\pm0.009$ & $0.907\pm0.009$ \\
			DNN+Graph-Attention & $0.903\pm0.109$ & $0.901\pm0.077$ & $0.683\pm0.002$ & $0.882\pm0.011$ & $0.904\pm0.009$ \\
			GCN+Graph-Attention & $\bf{0.927\pm0.013}$ & $\bf{0.915\pm0.009}$ & $\bf{0.690\pm0.017}$ & $\bf{0.889\pm0.008}$ & $\bf{0.909\pm0.005}$ \\
			\noalign{\smallskip}\hline
	\end{tabular}}
	
	\label{tab:ablation1}
\end{table}

The ablation experiment on the multi-instance learning method for graph structures is also carried out on the benchmark datasets for the multi-instance learning. The purpose of this experiment is to verify that the graph convolutional network and graph attention mechanism improve the performance of the multi-instance learning method, respectively. Table~\ref{tab:ablation1} compares the GCN-based MIL algorithm with some baselines to better appreciate the improvements provided by the architectural choices. The baselines are: i) the DNN-based MIL algorithm with the attention mechanism; ii) the DNN-based with the graph-attention mechanism; iii) the GNN-based MIL algorithm with the attention mechanism; vi) the GCN-based MIL algorithm with the attention mechanism. As can be seen from Table~\ref{tab:ablation1}, the experimental effect of the MIL algorithm based on DNN combined with our proposed graph-attention mechanism is better than an ordinary attention mechanism~\cite{ilse2018attention}. This verifies that the graph-attention mechanism fuses instances with complex topology structures to obtain better feature expressions of a bag. Moreover, the accuracy of a GCN-based MIL algorithm with an ordinary attention mechanism is higher than that of GNN-based MIL algorithm~\cite{tu2019multiple} with ordinary attention mechanism on MIL bechmark datasets. The accuracy of GNN-based MIL algorithm is higher than that of a DNN-based MIL algorithm. These verify that GCN fuses similar instances to obtain better feature expressions of each instance. In total, the two modules we have improved (the graph-attention mechanism and the graph convolutional network) improve the performance of multi-instance learning algorithms.

Automated detection of cancerous areas in $H\&E$ stained images is a clinically significant task. Current monitoring methods use pixel-level annotations; however, data preparation requires a lot of time for pathologists. Therefore, using a weak-label solution greatly reduces the workload of the pathologist. In the third experiment, we classify the breast cancer dataset and the colon cancer dataset to show that our proposed graph attention mechanism and method for constructing a bag graph structure are suitable for medical image analysis. For these two datasets, we use the trainable CNN framework proposed in~\cite{ilse2018attention} and a trainable ResNet18, whose parameters are pretrained on ImageNet, to represent the transformation function $f$. On these two datasets, we use Equations~\ref{eqn:3} and~\ref{eqn:2} to create the adjacent matrix $A$ of the graph, respectively.

\begin{table}
	\centering
	\caption{Results on the breast cancer dataset. Experiments are 5-fold cross-validation and an average ($\pm$ a standard error of the mean) is reported.}
	\resizebox{\textwidth}{!}{
		\begin{tabular}{llllll}
			\hline\noalign{\smallskip}
			Method & Accuracy & Precision & Recall & F-score & AUC\\
			\noalign{\smallskip}\hline\noalign{\smallskip}
			Ilse~\cite{ilse2018attention} & $0.755\pm0.016$ & $0.728\pm0.016$ & $0.731\pm0.042$ & $0.725\pm0.023$ & $0.799\pm0.020$  \\
			\noalign{\smallskip}\hline
			ours(Eq.~\ref{eqn:2}) & $0.808\pm0.070$ & $0.773\pm0.140$ & $0.800\pm0.110$ & $0.781\pm0.080$ & $0.816\pm0.070$\\ 
			\noalign{\smallskip}\hline
			ours(Eq.~\ref{eqn:3}) & $\bf{0.813\pm0.050}$ & $\bf{0.777\pm0.070}$ & $\bf{0.813\pm0.011}$ & $\bf{0.790\pm0.042}$ & $\bf{0.824\pm0.040}$\\
			\noalign{\smallskip}\hline
	\end{tabular}}
	\label{tab:2}
\end{table}

\begin{table}
	\centering
	\caption{Results on colon cancer dataset. Experiments are 5-fold cross-validation and an average ($\pm$ a standard error of the mean) is reported.}
	\resizebox{\textwidth}{!}{
		\begin{tabular}{llllll}
			\hline\noalign{\smallskip}
			Method & Accuracy & Precision & Recall & F-score & AUC\\
			\noalign{\smallskip}\hline\noalign{\smallskip}
			Ilse~\cite{ilse2018attention} & $0.898\pm0.020$ & $0.944\pm0.016$ & $0.851\pm0.035$ & $0.893\pm0.022$ & $0.968\pm0.010$  \\
			\noalign{\smallskip}\hline
			ours(Eq.~\ref{eqn:2}) & $0.931\pm0.040$ & $0.952\pm0.030$ & $0.892\pm0.040$ & $0.941\pm0.040$ & $0.971\pm0.038$ \\
			\noalign{\smallskip}\hline
			ours(Eq.~\ref{eqn:3}) & $\bf{0.940\pm0.020}$ & $\bf{0.970\pm0.070}$ & $\bf{0.900\pm0.030}$ & $\bf{0.954\pm0.050}$ & $\bf{0.978\pm0.028}$\\
			\noalign{\smallskip}\hline
	\end{tabular}}
	\label{tab:3}
\end{table}

We present the experimental results on the two datasets in Tables~\ref{tab:2} and~\ref{tab:3}, respectively. As shown in these two tables, our method outperforms the best MIL algorithm based on the attention mechanism proposed by Ilse et al.~\cite{ilse2018attention}. For image classification, we regard patches as instances and the image as a bag; each instance has an adjacent relationship on the original image. Our method does an excellent job of representing this relationship during image classification. Moreover, the attention mechanism in~\cite{tu2019multiple} is same as that in ~\cite{ilse2018attention}. Experimental results verify that  our proposed graph-attention mechainism is better. As can be seen from the last two rows of Tables~\ref{tab:2} and~\ref{tab:3}, experimental results of a graph built by structural relationships on the original image space domain outperform experimental results of a graph built by the similarities on the feature value domain. Therefore, our method of building a graph is suitable for image classification.

\begin{figure}[h]
	\centering
	\subfigure[]{
		\label{fig:a}
		\includegraphics[width=0.2\textwidth]{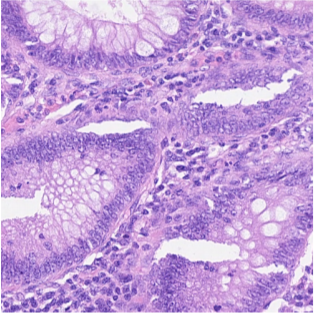}}
	\quad
	\subfigure[]{
		\label{fig:b}
		\includegraphics[width=0.2\textwidth]{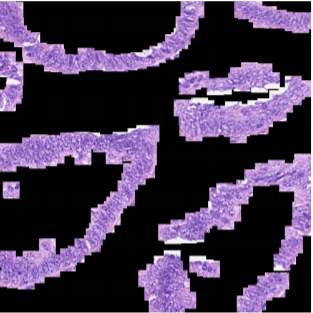}}
	\quad
	\subfigure[]{
		\label{fig:c}
		\includegraphics[width=0.2\textwidth]{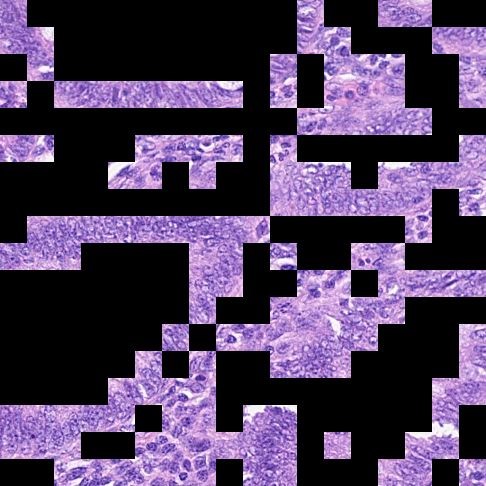}}
	\quad
	\subfigure[]{
		\label{fig:d}
		\includegraphics[width=0.2\textwidth]{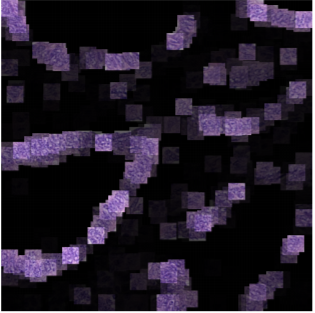}}
	\caption{(a) $H\&E$ stained histology image. (b) Ground truth: patches that belong to the class epithelial. (c) Heatmap of our method: every patch from (a) multiplied by its corresponding attention weight $\alpha$. (d) Heatmap of the attention-MIL method~\cite{ilse2018attention}. }
	\label{fig:rio}
\end{figure}

Finally, we use an image to verify that our method provides the ROI of a disease. As shown in Figure~\ref{fig:rio}, we show a histopathological image that is segmented into patches containing individual cells. We create a heatmap by multiplying the patch by the corresponding attention weight $\alpha$. Although we use image-level labels during training, there is a big overlap between the actual values in Figure~\ref{fig:b} and Figure~\ref{fig:c}. As shown in Figure~\ref{fig:d}, they manually process the image and input patches that overlap the region of interest, and then use the attention-MIL method~\cite{ilse2018attention} to obtain a heatmap. However, our method inputs all patches, which yields better results. The obtained results again verify that the graph-attention mechanism is effective and allows us to properly highlight ROIs.

\begin{table}[h]
	\footnotesize
	\centering
	\caption{Results on the IDRiD dataset. * represents the top two teams' results of the IDRiD challenge 2. Experiments are run five times, and an average is reported. }
	\begin{tabular}{llll}
		\hline\noalign{\smallskip}
		Method & DR(Ac) & DME(Ac) & DR+DME(Ac)\\
		\noalign{\smallskip}\hline\noalign{\smallskip}
		Safwan~\cite{safwan2018classification}& -- & -- & 0.476   \\
		VRT*~\cite{idrid}  & 0.592 &  0.8155 & 0.553 \\
		LzyUNCC*~\cite{idrid} & \bf{0.748} & 0.806 & 0.631  \\
		Wang~\cite{wang2018deep} & -- & -- & 0.660 \\
		Ma~\cite{ma2020pcfnet:} & -- & -- & 0.680\\
		\noalign{\smallskip}\hline
		ours & 0.728 & \bf{0.864} &\bf{0.689} \\
		\noalign{\smallskip}\hline
	\end{tabular}
	
	\label{tab:4}
\end{table}

\begin{table}[h]
	\centering
	\caption{Results on the ICIAR dataset. Experiments are 5-fold cross-validation and an average is reported.}
	\begin{tabular}{lll}
		\hline\noalign{\smallskip}
		Method(Ac) & 4-class & benign\ and\ invasive\\
		\noalign{\smallskip}\hline\noalign{\smallskip}
		Wang~\cite{wang2018classification} & 0.81 & -- \\
		Ilse~\cite{ilse2018attention} & 0.815 & 0.825 \\
		Combalia $\&$ Vilaplana~\cite{combalia2018monte} & -- & 0.847 \\
		\noalign{\smallskip}\hline
		ours & \bf{0.835} &\bf{0.905} \\
		\noalign{\smallskip}\hline
	\end{tabular}
	\label{tab:5}
\end{table}

To verify the third problem, we conduct experiments on the IDRiD and ICIAR datasets. In the experiments on these two image datasets, we use Equation~\eqref{eqn:3} to create the adjacent matrix $A$ of the graph. 

The original pixel information from the image and the adjacent relationships among patchs are very important for high-resolution image classification tasks~\cite{combalia2018monte}, and our approach is a good way to incorporate these. We present results on the two datasets in Tables~\ref{tab:4} and~\ref{tab:5}, respectively. As shown in Table~\ref{tab:4}, our approach outperforms the other methods, including the best non-MIL methods proposed by Wang et al.~\cite{wang2018deep} and Ma et al.~\cite{ma2020pcfnet:}. In these two methods, they first extract the feature vector with a pre-trained DenseNet121 or a PCF-DenseNet121 and then use the integrated algorithm (LightGBM) to obtain the final result. As can be seen in the last three lines of Table~\ref{tab:5}, our approach outperforms the other methods, including the best MIL algorithm proposed by Combalia $\&$ Vilaplana~\cite{combalia2018monte}. In this method, they focus on the most relevant regions of a high-resolution image using the Monte Carlo sampling strategy, and then iterates to get the most related patches. Judging from the results in the first and last row of Table~\ref{tab:5}, our method outperforms the non-MIL method proposed by Wang et al.~\cite{wang2018classification} wherein they use VGG16 with a pretraining strategy on similar datasets as the basic model to carry out the four classification tasks. In sum, compared to existing methods, our method is more suitable for high-resolution images classification.

\section{Conclusion}\label{sec:con}

In this paper, we present an end-to-end MIL algorithm based on GCN and a new graph-attention mechanism. We use structural relationships among instances in the image space domain to establish a graph, and then integrate this graph into GCN and the graph-attention mechanism to learn bag-embedding. By using the basic theorem of continuous function on the invariance of variable arrangement, we prove that the GCN-based MIL algorithm exhibits permutation invariance. We employ experimental analyses on five MIL benchmark datasets and four medical image datasets. Experimental results show that our method outperforms other methods, thus proving that our method is more suitable for high-resolution image classification. In addition, we have demonstrated that our approach presents ROI with attention weights to explain disease categories, which is important in many practical applications~\cite{xu2014weakly}.

In the future, we will explore the practical applications of our method in the detection of medical image lesions.


\bibliography{reference}

\begin{thebibliography}{10}
\expandafter\ifx\csname url\endcsname\relax
  \def\url#1{\texttt{#1}}\fi
\expandafter\ifx\csname urlprefix\endcsname\relax\def\urlprefix{URL }\fi
\expandafter\ifx\csname href\endcsname\relax
  \def\href#1#2{#2} \def\path#1{#1}\fi

\bibitem{zhou2009multi}
Z.-H. Zhou, Y.-Y. Sun, Y.-F. Li, Multi-instance learning by treating instances
  as non-iid samples, in: Proceedings of the 26th annual international
  conference on machine learning, ACM, 2009, pp. 1249--1256.

\bibitem{combalia2018monte}
M.~Combalia, V.~Vilaplana, {Monte-Carlo} sampling applied to multiple instance
  learning for histological image classification, in: Deep Learning in Medical
  Image Analysis and Multimodal Learning for Clinical Decision Support,
  Springer, 2018, pp. 274--281.

\bibitem{wang2018deep}
Y.~Wang, G.~A. Wang, W.~Fan, J.~Li, A deep learning based pipeline for image
  grading of diabetic retinopathy, in: International Conference on Smart
  Health, Springer, 2018, pp. 240--248.

\bibitem{xu2019camel}
G.~Xu, Z.~Song, Z.~Sun, C.~Ku, Z.~Yang, C.~Liu, S.~Wang, J.~Ma, W.~Xu, Camel: A
  weakly supervised learning framework for histopathology image segmentation,
  in: Proceedings of the IEEE International Conference on Computer Vision,
  2019, pp. 10682--10691.

\bibitem{hou2016patch}
L.~Hou, D.~Samaras, T.~M. Kurc, Y.~Gao, J.~E. Davis, J.~H. Saltz, Patch-based
  convolutional neural network for whole slide tissue image classification, in:
  Proceedings of the IEEE Conference on Computer Vision and Pattern
  Recognition, 2016, pp. 2424--2433.

\bibitem{cheplygina2015label}
V.~Cheplygina, L.~S{\o}rensen, D.~M. Tax, M.~de~Bruijne, M.~Loog, Label
  stability in multiple instance learning, in: International Conference on
  Medical Image Computing and Computer-Assisted Intervention, Springer, 2015,
  pp. 539--546.

\bibitem{andrews2003support}
S.~Andrews, I.~Tsochantaridis, T.~Hofmann, Support vector machines for
  multiple-instance learning, in: Advances in neural information processing
  systems, 2003, pp. 577--584.

\bibitem{chen2006miles}
Y.~Chen, J.~Bi, J.~Z. Wang, {MILES}: Multiple-instance learning via embedded
  instance selection, IEEE Transactions on Pattern Analysis and Machine
  Intelligence 28~(12) (2006) 1931--1947.

\bibitem{raykar2008bayesian}
V.~C. Raykar, B.~Krishnapuram, J.~Bi, M.~Dundar, R.~B. Rao, Bayesian multiple
  instance learning: automatic feature selection and inductive transfer., in:
  ICML, Vol.~8, 2008, pp. 808--815.

\bibitem{zhang2006adapting}
M.~Zhang, Z.~Zhou, Adapting rbf neural networks to multi-instance learning,
  Neural Processing Letters 23~(1) (2006) 1--26.

\bibitem{kandemir2015computer}
M.~Kandemir, F.~A. Hamprecht, Computer-aided diagnosis from weak supervision: A
  benchmarking study, Computerized medical imaging and graphics 42 (2015)
  44--50.

\bibitem{wang2018revisiting}
X.~Wang, Y.~Yan, P.~Tang, X.~Bai, W.~Liu, Revisiting multiple instance neural
  networks, Pattern Recognition 74 (2018) 15--24.

\bibitem{ilse2018attention}
M.~Ilse, J.~M. Tomczak, M.~Welling, Attention-based deep multiple instance
  learning, arXiv preprint arXiv:1802.04712.

\bibitem{xu2014weakly}
Y.~Xu, J.-Y. Zhu, I.~Eric, C.~Chang, M.~Lai, Z.~Tu, Weakly supervised
  histopathology cancer image segmentation and classification, Medical image
  analysis 18~(3) (2014) 591--604.

\bibitem{jia2017constrained}
Z.~Jia, X.~Huang, I.~Eric, C.~Chang, Y.~Xu, Constrained deep weak supervision
  for histopathology image segmentation, IEEE transactions on medical imaging
  36~(11) (2017) 2376--2388.

\bibitem{tu2019multiple}
M.~Tu, J.~Huang, X.~He, B.~Zhou, Multiple instance learning with graph neural
  networks, arXiv preprint arXiv:1906.04881.

\bibitem{zhang2011multiple}
D.~Zhang, Y.~Liu, L.~Si, J.~Zhang, R.~D. Lawrence, Multiple instance learning
  on structured data, in: Advances in Neural Information Processing Systems,
  2011, pp. 145--153.

\bibitem{zhou2007relation}
Z.-H. Zhou, J.-M. Xu, On the relation between multi-instance learning and
  semi-supervised learning, in: Proceedings of the 24th international
  conference on Machine learning, ACM, 2007, pp. 1167--1174.

\bibitem{ma2020pcfnet:}
Y.~Ma, Y.~Luo, Z.~Yang, Pcfnet: Deep neural network with predefined
  convolutional filters, Neurocomputing 382 (2020) 32--39.

\bibitem{zaheer2017deep}
M.~Zaheer, S.~Kottur, S.~Ravanbakhsh, B.~Poczos, R.~R. Salakhutdinov, A.~J.
  Smola, Deep sets, in: Advances in neural information processing systems,
  2017, pp. 3391--3401.

\bibitem{kipf2016semi}
T.~N. Kipf, M.~Welling, Semi-supervised classification with graph convolutional
  networks, arXiv preprint arXiv:1609.02907.

\bibitem{wang2018classification}
Z.~Wang, N.~Dong, W.~Dai, S.~D. Rosario, E.~P. Xing, Classification of breast
  cancer histopathological images using convolutional neural networks with
  hierarchical loss and global pooling, in: International Conference Image
  Analysis and Recognition, Springer, 2018, pp. 745--753.

\bibitem{idrid}
P.~Porwal, S.~Pachade, R.~Kamble, M.~Kokare, G.~Deshmukh, V.~Sahasrabuddhe,
  F.~Meriaudeau, Indian diabetic retinopathy image dataset (idrid): a database
  for diabetic retinopathy screening research, Data 3~(3) (2018) 25.

\bibitem{gartner2002multi}
T.~G{\"a}rtner, P.~A. Flach, A.~Kowalczyk, A.~J. Smola, Multi-instance kernels,
  in: ICML, Vol.~2, 2002, p.~7.

\bibitem{zhang2002dd}
Q.~Zhang, S.~A. Goldman, {EM-DD}: An improved multiple-instance learning
  technique, in: Advances in neural information processing systems, 2002, pp.
  1073--1080.

\bibitem{wei2016scalable}
X.-S. Wei, J.~Wu, Z.-H. Zhou, Scalable algorithms for multi-instance learning,
  IEEE transactions on neural networks and learning systems 28~(4) (2016)
  975--987.

\bibitem{safwan2018classification}
M.~Safwan, S.~S. Chennamsetty, A.~Kori, V.~Alex, G.~Krishnamurthi,
  Classification of breast cancer and grading of diabetic retinopathy \&
  macular edema using ensemble of pre-trained convolutional neural networks.

\end{thebibliography}

\end{document}